\documentclass[a4paper, 10pt, conference]{ieeeconf}      % Use this line for a4 paper

\IEEEoverridecommandlockouts                              % This command is only needed if 
\usepackage{graphics} % for pdf, bitmapped graphics files
\usepackage{epsfig} % for postscript graphics files
\usepackage{times} % assumes new font selection scheme installed
\usepackage{amsmath} % assumes amsmath package installed
\usepackage{amssymb}  % assumes amsmath package installed

\usepackage[nocomma]{optidef}
\usepackage{algpseudocode}
\usepackage{mathtools}
\usepackage{bm}
\usepackage[ruled,vlined]{algorithm2e}

\usepackage{float}
\usepackage{subfig}
\usepackage{multicol}
\usepackage{multirow}
\usepackage{makecell}
\usepackage{booktabs}
\usepackage{adjustbox}

\usepackage[utf8]{inputenc}
\usepackage[english]{babel}
\usepackage{hyperref}
\usepackage{booktabs}
\usepackage{multirow}
\usepackage{makecell}
\usepackage{stfloats}
\usepackage{float}
\let\labelindent\relax
\usepackage{enumitem}
\usepackage{pgfplots}
\hypersetup{
    colorlinks=true,
    linkcolor=blue,
    filecolor=magenta,      
    urlcolor=cyan,
    pdftitle={Safe Area Navigation of Mobile Robots via Model Predictive Control},
    pdfpagemode=FullScreen,
}

\title{\LARGE \bf
Collision-Free Navigation of Mobile Robots via Quadtree-Based Model Predictive Control
}

\author{Osama Al Sheikh Ali, Sotiris Koutsoftas, Ze Zhang, Knut Åkesson, Emmanuel Dean
\thanks{This work was partially supported by the Chalmers Production Area of Advance (AoA) project.}% <-this % stops a space
\thanks{All authors are with Electrical Engineering, Chalmers University of Technology, 41296 Gothenburg, Sweden. Contact:
        {\tt\small zhze@chalmers.se}}%
}

\begin{document}

\maketitle
\thispagestyle{empty}
\pagestyle{empty}

%%%%%%%%%%%%%%%%%%%%%%%%%%%%%%%%%%%%%%%%%%%%%%%%%%%%%%%%%%%%%%%%%%%%%%%%%%%%%%%%
\begin{abstract}
This paper presents an integrated navigation framework for Autonomous Mobile Robots (AMRs) that unifies environment representation, trajectory generation, and Model Predictive Control (MPC). The proposed approach incorporates a quadtree-based method to generate structured, axis-aligned collision-free regions from occupancy maps. These regions serve as both a basis for developing safe corridors and as linear constraints within the MPC formulation, enabling efficient and reliable navigation without requiring direct obstacle encoding. The complete pipeline combines safe-area extraction, connectivity graph construction, trajectory generation, and B-spline smoothing into one coherent system. Experimental results demonstrate consistent success and superior performance compared to baseline approaches across complex environments.
\end{abstract}
%%%%%%%%%%%%%%%%%%%%%%%%%%%%%%%%%%%%%%%%%%%%%%%%%%%%%%%%%%%%%%%%%%%%%%%%%%%%%%%%

\section{Introduction}

Autonomous Mobile Robots (AMRs) are becoming pivotal across industrial domains, requiring precision, flexibility, and adaptability \cite{Alatise_2020_AMRs}.
Safe and efficient navigation in such environments typically mandates integrating three core components: global path planning, trajectory generation, and control for trajectory tracking \cite{amrintro_2011_siegwart}, which are usually decoupled. A global planner, such as A* or RRT, generates a geometric path without accounting for robot kinematics and dynamics, while a local controller attempts to track it \cite{Berlin}. This separation can lead to violations of safety or dynamic feasibility constraints \cite{katayama2023mpc, Haffemayer_2024_HardConstraints}, as the geometric path may require maneuvers exceeding the robot’s acceleration, steering, or obstacle avoidance capabilities when executed in real environments.

To overcome this decoupling between planning and control, recent work has adopted Model Predictive Control (MPC), which unifies both processes through receding-horizon optimization under dynamic and safety constraints~\cite{Stella_2017_panoc, rakovic2021exclusion}. However, representing obstacles as hard constraints may yield non-convex conditions that hinder real-time performance and robustness~\cite{Haffemayer_2024_HardConstraints, zhang_2025_multimodalmpc}. Conventional MPC explicitly encodes convex polygonal obstacles as constraints, causing computational spikes or infeasibility in cluttered environments. Alternative variants mitigate this issue using penalty methods or by tracking precomputed paths from global planners~\cite{Berlin, penaltyNMPC}, but remain sensitive to parameter tuning and prone to local minima. Deep Reinforcement Learning (DRL) offers a data-driven alternative~\cite{glanois2024survey, DRL}, yet its black-box nature limits interpretability and trustworthiness in safety-critical applications.
%%%%%%%%%%%%%%%%%%%%%%%%%%%%%%%%%%%%%%%%%%%%%%%%%%%%%%%%%%%%%%%%%%%%%%%%%

Several studies leverage structured environment representations to improve trajectory planning efficiency. In~\cite{chen2016collisionfree}, safe flight corridors are constructed using octree decomposition and generate trajectories via quadratic programming, while requiring iterative extrema checking to satisfy corridor constraints. It is extended to multi-agent coordination through relative corridors based on Bernstein polynomial properties~\cite{park2019safecorridor}. In~\cite{Brüdigam_2020_gridmpc}, a grid-based stochastic MPC is formulated that thresholds probabilistic occupancy maps and computes convex hulls for constraint enforcement.
%%%%%%%%%%%%%%%%%%%%%%%%%%%%%%%%%%%%%%%%%%%%%%%%%%%%%%%%%%%%%%%%%%%%%%%%%

\begin{figure}[t]
   \centering
   \includegraphics[width=\linewidth]{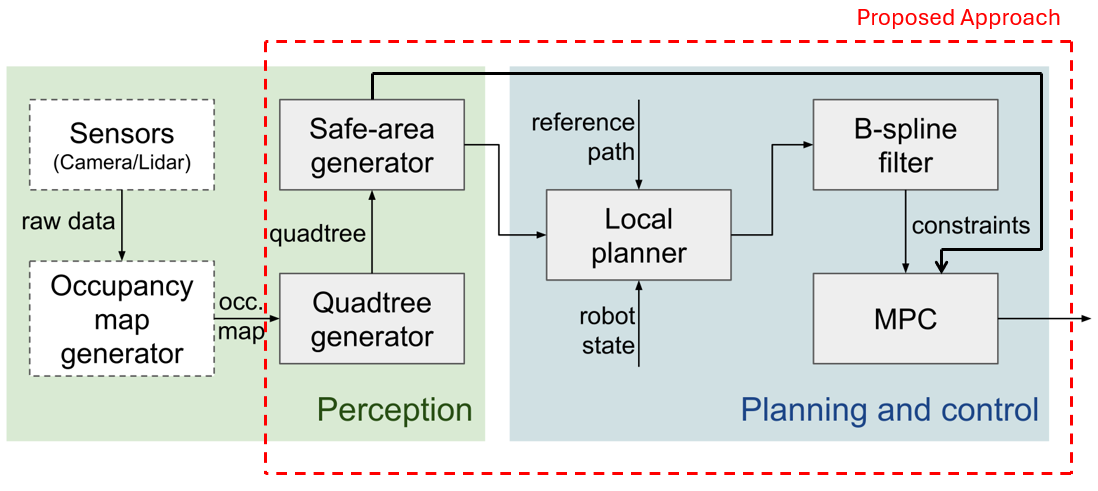}
   % \caption{The proposed framework for Perception, Planning and Control through Sensors, Occupancy Map Generator, Quadtree Generator, Safe-Area Generator, Local Path Planner, B-Spline Filtering, constraints and MPC.}
   \vspace{-7mm}
   \caption{The proposed framework integrates quadtree-based safe-area generation from occupancy grid map perception with local planning and control.}
   \vspace{-4mm}
   \label{fig:sampc}
\end{figure}

In this work, the proposed quadtree-based approach enables one-shot convex constraint generation through hierarchical spatial decomposition and efficient region merging, which doesn't require iterative constraint refinement ~\cite{chen2016collisionfree}, or relies on probabilistic thresholding~\cite{Brüdigam_2020_gridmpc}. This design directly addresses the persistent challenge of achieving feasible navigation with formal safety guarantees in environments containing non-convex obstacles. In general, the framework integrates perception, mapping, trajectory generation, and control within an MPC formulation as demonstrated in Fig.~\ref{fig:sampc}. The occupancy map is decomposed into axis-aligned safe regions, which are subsequently merged into larger convex areas that serve a dual purpose: they define a connectivity-aware reference trajectory and are simultaneously encoded as linear inequality constraints in the MPC problem. This unified use of precomputed safe areas for both path generation and constraint enforcement eliminates the need to model obstacles directly, transforming the inherently non-convex obstacle-avoidance task into a convex optimization problem. As a result, the approach guarantees recursive feasibility, improves computational stability, and enhances robustness in complex, cluttered environments. 

The main innovations of this work are threefold: (1) a safe-area MPC framework for dynamic feasibility and safety by tightly coupling all navigation components; (2) a quadtree-derived convex constraint embedding that makes obstacle avoidance tractable; and (3) a dual-layer MPC cost structure that combines hard safety constraints with soft penalty margins to improve resilience against modeling uncertainties.

\section{PRELIMINARIES}

For mobile robots, MPC utilizes their motion model to anticipate future states and achieve certain objectives while satisfying system and safety constraints.
At each control cycle, given the horizon $N$, it solves:
\begin{align}
    \min_{\bm{u}_{0:N-1}} & \sum_{k=0}^{N-1} \ell(\bm{x}_k, \bm{u}_k) + \ell_N(\bm{x}_N), \\
    \text{s.t.} \quad %& \bm{x}_0 = \bm{x}(t), \\
    & \bm{x}_{k+1} = f(\bm{x}_k, \bm{u}_k), \\
    & \bm{x}_k \in \mathcal{X},  \\
    & \bm{u}_k \in \mathcal{U},
\end{align}
where \( \bm{x}_k \) and \( \bm{u}_k \) are the robot state and input at time step $k$, \( f(\cdot) \) is the robot’s motion model, \( \ell(\bm{x}_k, \bm{u}_k) \) is the stage cost, and \( \ell_N(\bm{x}_N) \) is a terminal cost. The constraints \( \mathcal{X} \) and \( \mathcal{U} \) encode physical and safety limits.
Only the first input $\bm{u}_0$ is applied at each step (receding horizon), allowing MPC to adapt to changing conditions while optimising future behaviour.

\section{METHODOLOGY}
\subsection{Quadtree-based Safe Area Generation}
\label{sec:safearea}

\begin{figure*}[t]
    \centering
    \includegraphics[width=0.9\textwidth]{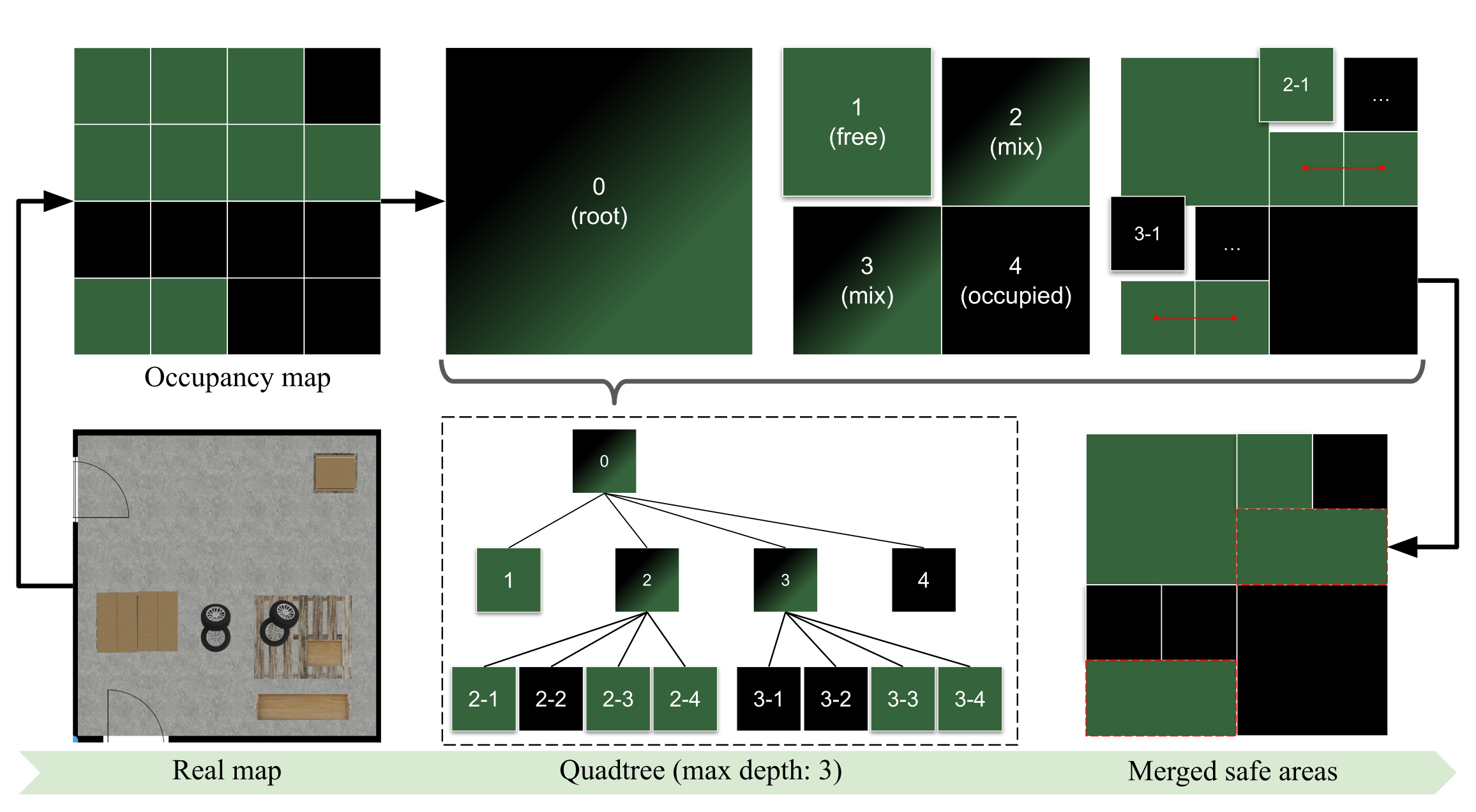}
    \vspace{-3mm}
    \caption{Illustration of the safe area generation: From the real-world map to occupancy map generation, quadtree decomposition (nodes are indexed with different colors representing their states), and merged safe areas enabling efficient trajectory planning.}
    \vspace{-3mm}
    \label{fig:safe_area_generation}
\end{figure*}

Occupancy grids divide the environment into uniform cells indicating free or occupied space~\cite{amrintro_2011_siegwart}, enabling reliable segmentation and collision checking. Although this uniform representation is effective for basic mapping and navigation, it becomes inefficient in large or unevenly structured environments due to redundant data in homogeneous regions. To overcome this limitation, quadtrees extend the occupancy grid concept by hierarchically subdividing non-uniform regions into quadrants while merging homogeneous areas, providing a more compact and flexible spatial representation. By recursively partitioning the occupancy map (Fig.~\ref{fig:safe_area_generation}), the quadtree clusters contiguous free spaces into coherent regions and continues this hierarchical subdivision until each region reaches homogeneity or the minimum resolution $\delta_{\min}$, yielding leaf nodes that represent free or occupied areas. This hierarchical representation preserves spatial relationships between navigable regions and inherently compresses map data while maintaining adjacency information between free regions. 
The overall quadtree-based safe area generation procedure is summarised in Algorithm~\ref{alg:safe_area_generation}.

\begin{algorithm}[t]
\SetAlgoLined

\KwIn{Occupancy map of the environment, desired minimum cell size}
\KwOut{Quadtree representation with safe area connectivity graph}

% Step 1: Create root node covering the entire map
Initialize \texttt{QuadTreeNode} for the full map.

% Step 2: Recursively subdivide mixed regions
\texttt{build\_quadtree(node)}:
\begin{enumerate}[leftmargin=*]
  \item If the region is free or occupied, mark as leaf.
  \item If free, add to the connectivity graph.
  \item Otherwise, split into quadrants, repeat for each child.
  \item Then, call \texttt{update\_connectivity(node)}.
\end{enumerate}
\vspace{1mm}

% Step 3: Link neighboring free regions
\texttt{update\_connectivity(node)}:
\begin{enumerate}[leftmargin=*]
  \item For each adjacent free region sharing a full edge, \\  
        add an edge in the connectivity graph weighted by centre distance.
\end{enumerate}
\vspace{1mm}

% Step 4: Merge adjacent free regions into larger safe areas
\texttt{merge\_adjacent\_nodes()}:
\begin{enumerate}[leftmargin=*]
  \item For each free region, check its neighbors.
  \item If two regions can merge into a valid rectangle, combine them.
  \item Update connectivity and repeat until no more \\ merges are possible.
\end{enumerate}

\caption{Quadtree-based safe area generation and merging procedure.}
\label{alg:safe_area_generation}
\end{algorithm}

\subsection{Safe Area Merging and Connectivity}
While this fine-grained representation captures local spatial structure accurately, it introduces significant challenges for downstream planning. In particular, the large number of small and potentially disconnected safe regions leads to a dense and complex connectivity graph. This not only increases the computational cost of graph construction but also results in longer planning times, as the search space becomes cluttered with redundant transitions.

For motion planners, this fragmentation means more region boundaries must be considered, each potentially introducing additional constraints during optimisation. As a result, the planner may produce conservative or suboptimal trajectories due to restricted manoeuvring flexibility.
To address this, merging areas reduces data redundancy and enables efficient multi-scale and variable-resolution mapping~\cite{SENOUSSI1994387,6629556}. The \texttt{merge\_adjacent\_nodes} function is applied, which consolidates neighbouring rectangular free cells into larger convex regions. Two cells are merged if they share a complete edge and their union forms a valid rectangle. This reduces region fragmentation, resulting in fewer transitions and a more tractable planning space for MPC. Following the merging process, the \texttt{update\_connectivity} function reconstructs the connectivity graph by linking adjacent free regions. This yields a simplified topological map where edges represent direct, obstacle-free traversals between safe areas. The effectiveness of this process is illustrated in Fig.~\ref{fig:safe_area_generation}, where the initial dense cell map is transformed into a more coherent and navigable environment.

\subsection{Trajectory Generation through collision-free areas}
%\label{sec:trajgeneration}
To generate a smooth, collision-free trajectory based on the quadtree-derived safe corridors, the process comprises three stages: (i) discrete waypoint selection, (ii) interpolation of waypoints into a dense reference path, and (iii) reference trajectory generation via B-spline smoothing.

\begin{figure}[t]
    \centering
    \includegraphics[width=.9\linewidth]{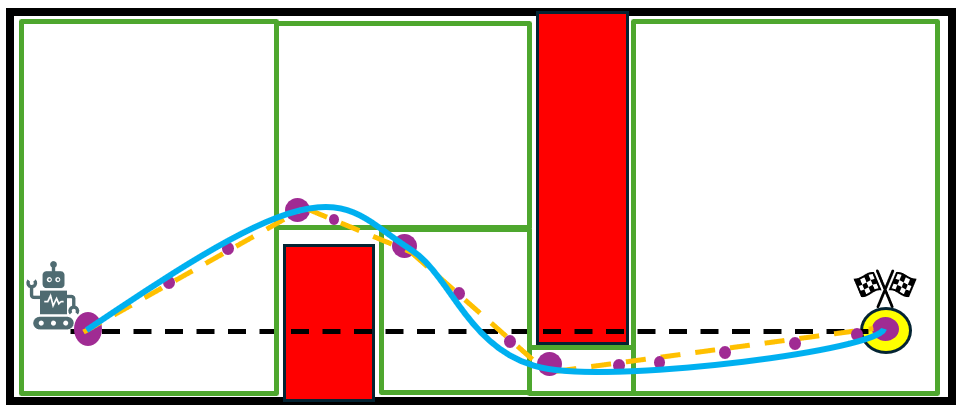}
    \caption{Trajectory generation. Green rectangles are safe regions, and red blocks are obstacles. The dark dashed line is the original reference path. Large purple points are waypoints from greedy search, and small ones are interpolated. The yellow polyline forms the new reference path, and the blue curve is the B-spline-smoothed trajectory.}
    \vspace{-2mm}
    \label{fig:traj_gen}
\end{figure}

\textbf{Waypoints generation}:
The first stage constructs a sequence of waypoints that connect the start and goal through adjacent safe polygons. The sequence starts with the robot’s current position $\bm{p}_0$ and ends at the goal position $\bm{p}_g$. The system identifies the safe polygons containing the start and goal (or the nearest ones if outside). If these polygons are different, the goal point is projected to the safe area adjacent to the start polygon. Specifically, this procedure identifies the closest point on the polygon boundary to the input goal and computes a small inward offset along the estimated normal vector to ensure the projection lies strictly inside the safe region. Fig. \ref{fig:traj_gen} illustrates the waypoints generation procedure, with large purple-coloured points marking the start and projected goal waypoints.

\textbf{Waypoints interpolation}:
Given a sequence of waypoints $\{\bm{w}_i\}_{i=0}^M$ starting at \(\bm{w}_0\!=\!\bm{p}_0\) and ending at \(\bm{w}_M\!=\!\bm{p}_g\), the path is parametrised by its cumulative arc‑length. 
Starting with \(s_0 \!=\! 0\) and incrementally adding segment lengths \(s_i \!=\! s_{i-1} + \lVert \bm{w}_i - \bm{w}_{i-1} \rVert\), the total length is $s_M$. For any $s\in[0,\,s_M]$, the segment index $i$ is located such that $s \in [s_i, s_{i+1})$. Then, a normalised interpolation ratio can be defined, and the interpolated point can be computed:
\begin{align}
    \bm{w}'(s) &= (1-\alpha(s))\,\bm{w}_i + \alpha(s)\,\bm{w}_{i+1}, \label{eq:interp} \\
    \text{where}\quad\alpha(s) &= (s-s_i)/(s_{i+1}-s_i)
\end{align}
Note that \(\bm{w}'(s)\) always lies inside the convex hull of \(\bm{w}_i\) and \(\bm{w}_{i+1}\). Since each segment is contained in the corridor, \(\bm{w}'(s) \in\mathrm{conv}(\bm{w}_i,\,\bm{w}_{i+1})\subseteq\mathcal{S}\), guarantees collision‑free interpolation. The arc‑length parametrisation also provides a well‑defined domain \([0,s_M]\) for sampling, using uniform steps in \(s\) ensures that interpolated points are distributed evenly along the path regardless of waypoint spacing. 

\textbf{B–spline smoothing}:
To obtain a continuously differentiable trajectory with reduced curvature and to ensure compliance with the robot’s dynamics, a cubic B–spline \(\bm{S}(\mu)\) is fitted to the densified reference path. With \(\{\bm{p}_j\}_{j=1}^M\) denoting the interpolated positions sampled uniformly along the arc‑length parameter \(s\). The spline control points are determined by minimising the regularised least–squares cost:
\begin{equation}
    J = \sum_{j=1}^{M} \Vert \bm{S}(\mu_j) - \bm{p}_j\Vert^2 + \lambda\int_{0}^{1}\Vert \bm{S}''(\mu)\Vert^2\,\mathrm{d}\mu,
    \label{eq:bsplineobj}
\end{equation}
where \(\mu\in[0,1]\) is the spline parameter and \(\lambda>0\) weights the smoothness term.  The first term enforces closeness to the reference path, while the second penalises high curvature. Once the B–spline is computed, its samples form the reference trajectory are used by the MPC.

\subsection{Safe Area Constraints and Safety Guarantee}

As a primary contribution, the incorporation of safety-area linear inequality constraints into the MPC formulation enforces navigation within the designated safe areas. Each rectangular safe region is represented as a set of four linear inequalities, which are imposed as hard constraints.
These are expressed in Eq.~\eqref{eq:safety_constraints}, where \( \bm{A}_{\text{safe}, k} \) and \( \bm{b}_{\text{safe}, k} \) define the spatial boundaries at step \( k \), and \( \bm{X}(t) \) is the system state:
\begin{align}
    \bm{A}_{\text{safe}, k} \, \bm{X}(t) \leq \bm{b}_{\text{safe}, k}.
    \label{eq:safety_constraints}
\end{align}
To improve resilience against modelling errors or unexpected dynamics, a soft constraint in the form of a penalty term \( J_\text{SA} \) is introduced. This term quantifies deviations from the safe regions, encouraging trajectories that remain within convex boundaries defined as:
\begin{align}
    H_{n,m} = \{ \bm{x} \in \mathbb{R}^2 : b_{n,m} - a_{n,m}^\top \bm{x} > 0 \},
\end{align}
where \( h_{n,m}(\bm{x}) = b_{n,m} - a_{n,m}^\top \bm{x} \) evaluates the robot’s relative position to each boundary.
Safety is guaranteed when:
\begin{align}
\prod_{n=1}^{N} \sum_{m=1}^{4} \left[ h_{n,m}(\bm{x}) \right]^2_{-} = 0,
\end{align}
and is penalized when violated via the smooth cost function:
\begin{align}
J_{\text{SA}}(\bm{x}) = \prod_{n=1}^{N} \sum_{m=1}^{4} \left[ q_{\text{SA}}(n) \cdot h_{n,m}(\bm{x}) \right]^2_{-},
\end{align}
where $[h]_{-} = \min(0, h)$, and \( q_{\text{SA}}(n) \) controls the priority of each safe area.

By embedding $J_{\text{SA}}$ into the MPC cost, the planner not only avoids unsafe regions but also actively penalizes proximity to their boundaries. This dual-layer formulation of hard linear constraints and soft penalties represents a distinctive integration of safety considerations into MPC, ensuring that all planned actions remain within validated safe zones.
Crucially, the inclusion of $J_{\text{SA}}$ acts as a safety guarantee within the MPC optimisation. Even in uncertainty or dense environments, the generated control actions strictly adhere to safety boundaries, enabling the robot to operate with full assurance of collision-free behaviour at every step.

\subsection{Full MPC Formulation}

For the robot's discrete state space, at time step $k$, the state vector \( \bm{x}_k = [x_k, y_k, \theta_k]^\top \) represents the robot's position \( (x_k, y_k) \) and its heading angle \( \theta_k \) in Cartesian space, while the control input \( \bm{u}_k = [v_k, \omega_k]^\top \) consists of the linear speed \( v_k \) and angular velocity \( \omega_k \). The motion model is the unicycle model, assuming no wheel slip:
\begin{equation}
f(\bm{x}_{k}, \bm{u}_k) =
\begin{bmatrix}
x_k + v_k \cos(\theta_k) \Delta t \\
y_k + v_k \sin(\theta_k) \Delta t \\
\theta_k + \omega_k \Delta ts
\end{bmatrix},
\label{eq:robot_kinematics}
\end{equation}
\noindent where \( \Delta t \) is the sampling time. The key contribution here is the incorporation of quadtree-derived safe areas directly into the MPC constraints. Each rectangular safe region from the quadtree decomposition is encoded as four linear inequalities, transforming the non-convex collision avoidance problem into a set of convex constraints. The complete MPC problem incorporating safe area navigation is formulated as:
\begin{align}
    \min_{\bm{u}_{0:N-1}} & \quad J_N + \sum_{k=0}^{N-1} \left[ J_R(k) + J_{SA}(\bm{x}_k) \right] \label{eq:MPC_equation} \\
    \text{s.t.} %& \quad \bm{x}_0 = \bar{\bm{x}}, \nonumber \\
    & \quad \bm{x}_{k+1} = f(\bm{x}_k, \bm{u}_k), \quad k = 0, 1, \ldots, N-1, \nonumber \\
    & \quad \bm{u}_k \in [\bm{u}_{\text{min}}, \bm{u}_{\text{max}}], \quad k = 0, 1, \ldots, N-1, \nonumber \\
    & \quad \frac{\Delta \bm{u}_k}{\Delta t} \in [\dot{\bm{u}}_{\text{min}}, \dot{\bm{u}}_{\text{max}}], \quad k = 0, 1, \ldots, N-1, \nonumber \\
    & \quad \underbrace{\bm{A}_{\text{safe}, k} \, \bm{x}_k \leq \bm{b}_{\text{safe}, k}}_{\text{Quadtree-derived constraints}}, \quad k = 0, 1, \ldots, N, \nonumber
\end{align}
where $\bm{x}_{0}$ is the current state, the terminal cost is $J_N = \left\| \bm{x}_N - \tilde{\bm{x}}_N \right\|^2_{Q_N}$ with the penalty value $Q_N$.
The quadtree information is embedded through the constraint, \(\bm{A}_{\text{safe}, k} \, \bm{x}_k \leq \bm{b}_{\text{safe}, k}\), where \(\bm{A}_{\text{safe}, k} \in \mathbb{R}^{4 \times 2}\) contains the normal vectors of the four edges of the rectangular safe area at prediction step $k$, and \(\bm{b}_{\text{safe}, k} \in \mathbb{R}^{4}\) contains the corresponding offset values, which parameters are extracted from the quadtree leaf nodes identified as collision-free during the safe area generation phase. 
The running cost \(J_R(k)\) is given by:
\begin{equation}
\small
J_R(k) = \| \bm{x}_k - \tilde{\bm{x}}_k \|^2_{Q_x} + \| \bm{u}_k - \tilde{\bm{u}}_k \|^2_{Q_u} + \| \bm{u}_k - \bm{u}_{k-1} \|^2_{Q_a},
\end{equation}
which balances trajectory tracking, control accuracy, and smoothness, where \( \| \bm{x}_k - \tilde{\bm{x}}_k \|^2_{Q_x} \) minimises the deviation from the desired state, \( \| \bm{u}_k - \tilde{\bm{u}}_k \|^2_{Q_u} \) penalises control signal deviations, and \( \| \bm{u}_k - \bm{u}_{k-1} \|^2_{Q_a} \) ensures smooth transitions between control inputs. The weighting matrices \( Q_x \), \( Q_u \), and \( Q_a \) tune the importance of each term. The control input \(u_k\) is constrained within a specified range \([\bm{u}_{\text{min}}, \bm{u}_{\text{max}}]\), and its rate of change \(\frac{\Delta \bm{u}_k}{\Delta t}\) is also bounded by \([\dot{\bm{u}}_{\text{min}}, \dot{\bm{u}}_{\text{max}}]\). Finally, the penalty constraint, \(\bm{A}_{\text{safe}, k} \, \bm{X}(t) \leq \bm{b}_{\text{safe}, k}\), keeps the robot within predefined safe limits.

\section{Implementation and Results}
The proposed safe-area MPC framework is evaluated under diverse scenarios. All experiments were conducted with an AMD Ryzen 7 5800H CPU, with randomized initial orientation to assess robustness. The MPC problem is solved by the PANOC solver \cite{Stella_2017_panoc} for real-time performance, and its weighting matrices were selected to prioritize adherence to the reference path while maintaining smooth movement. 
In this work, all experiments are in static environments, so the quadtree is computed offline in advance.

\subsection{Evaluation Again Explicit Obstacle Avoidance}
To evaluate the effectiveness of the proposed method, we compare it with a conventional MPC baseline that represents established trajectory planners, such as in \cite{katayama2023mpc, MPC1, MPC2}. This baseline follows a nonlinear MPC formulation that plans motion over a finite horizon while minimizing a trajectory cost and satisfying collision avoidance constraints. The baseline uses the same kinematic model and tuning parameters as the proposed SA-MPC. Obstacles are explicitly modelled as convex polygons and, in this implementation, are incorporated into the MPC formulation through terms in the cost function rather than as hard feasibility constraints \cite{Hattori2006}.

\begin{figure}[t]
    \centering
    \includegraphics[width=.9\linewidth]{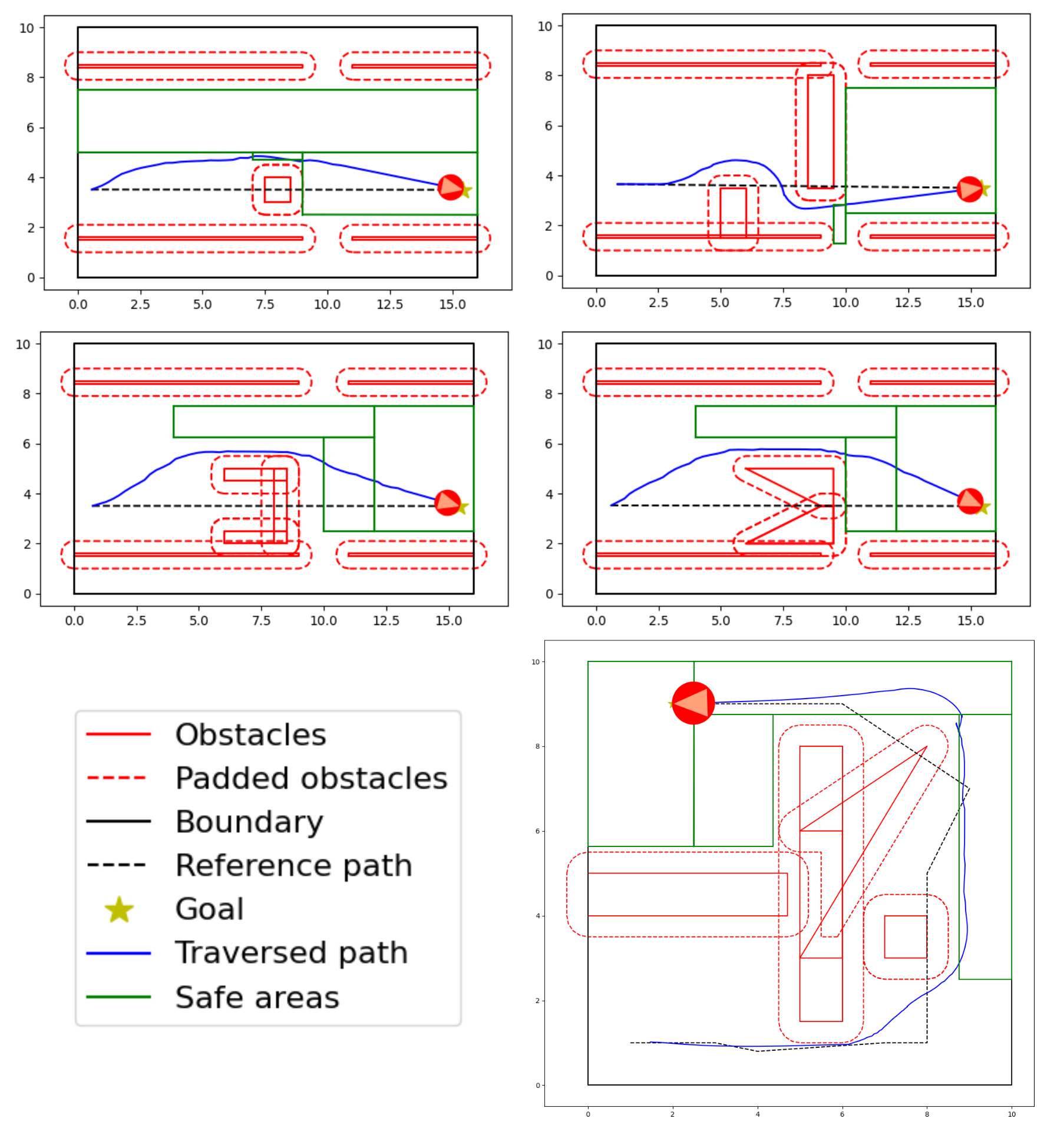}
    \vspace{-2mm} % tighten vertical space
    \caption{SA-MPC navigation across test scenarios with quadtree-derived safe areas (green)}
    \vspace{-2mm}

    \label{fig:mpc_scenarios}
\end{figure}

\begin{figure}[t]
    \centering
    \includegraphics[width=.9\linewidth]{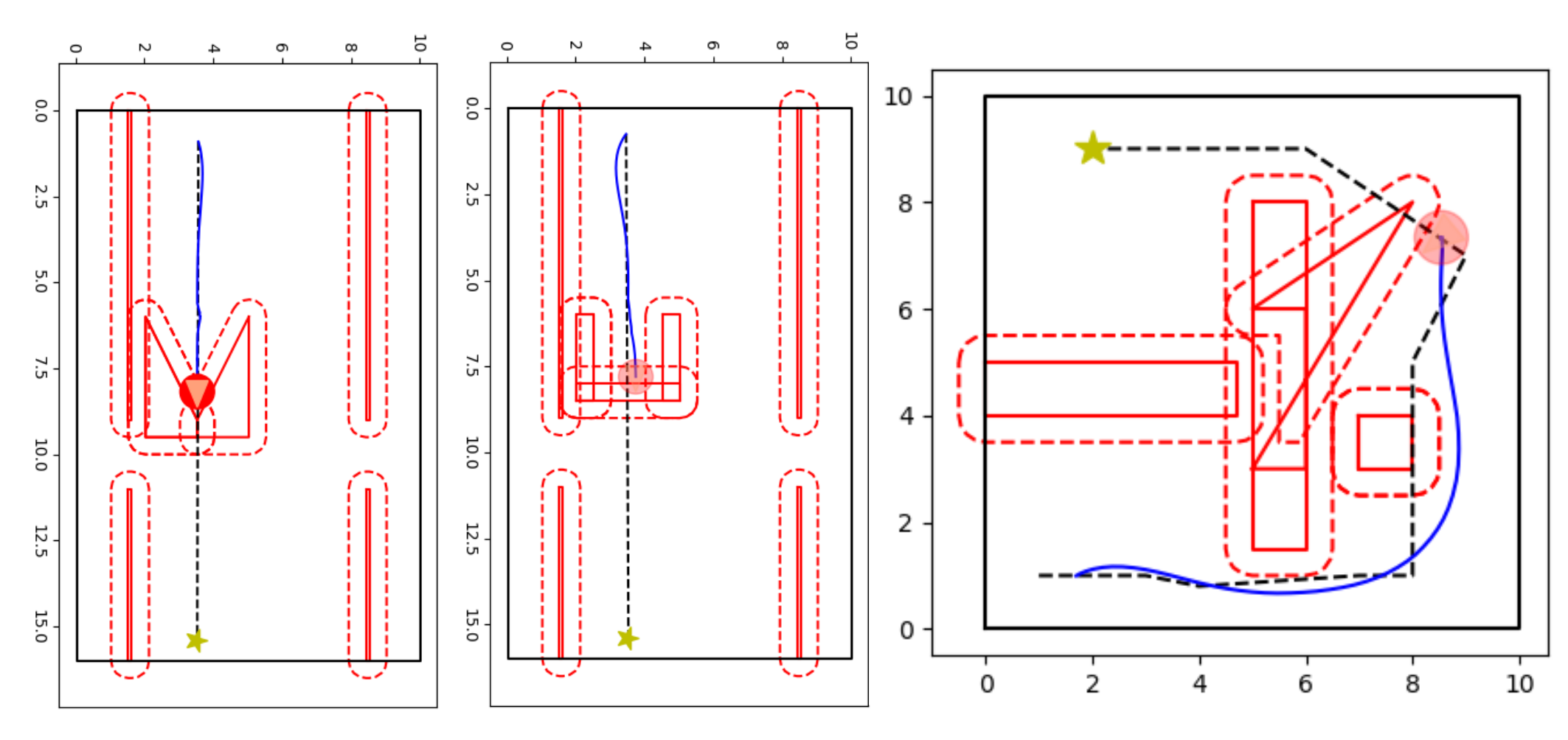}
    \caption{Baseline MPC performance showing failures in non-convex and complex scenarios}
    \label{fig:ref_mpc_scenarios}
\end{figure}

\pgfplotsset{compat=1.18}
\usetikzlibrary{patterns}

\begin{figure}[t]
  \centering
  % Data table
  \pgfplotstableread[row sep=\\,col sep=&]{
  scenario        & MPC & DQN & DWA & SA\\
  Single rec.     & 100 &  84 & 100 & 100\\
  Double rec.     &   2 & 100 & 100 & 100\\
  U-shape         &   2 &  94 &   2 & 100\\
  V-shape         &   2 &   2 &   2 & 100\\
  }\srdata

  \begin{tikzpicture}
    \begin{axis}[
      ybar,
      width=\linewidth,
      height=0.5\linewidth,
      ymin=0, ymax=100,
      ymajorgrids,
      grid style={densely dotted},
      ylabel={Success rate (\%)},
      symbolic x coords={Single rec., Double rec., U-shape, V-shape},
      xtick=data,
      xticklabel style={font=\scriptsize, rotate=15, anchor=east, align=center},
      bar width=6pt,
      enlarge x limits=0.18,
      legend style={at={(0.5,1.02)}, anchor=south, legend columns=4,
                    /tikz/every even column/.style={column sep=6pt}},
      legend image code/.code={\draw[#1] (0cm,-0.05cm) rectangle (0.27cm,0.15cm);},
      ytick={0,20,40,60,80,100},
      tick style={black},
    ]

    % Distinct colors & patterns (each defined in its own \addplot)
    \addplot+[ybar, bar shift=-0.3cm, fill=blue!60,   draw=black] 
      table[x=scenario,y=MPC]{\srdata};
    \addplot+[ybar, bar shift=-0.1cm, fill=cyan!60,   draw=black] 
      table[x=scenario,y=DQN]{\srdata};
    \addplot+[ybar, bar shift=0.1cm,  fill=orange!70, draw=black] 
      table[x=scenario,y=DWA]{\srdata};
    \addplot+[ybar, bar shift=0.3cm,  fill=red!60,    draw=black] 
      table[x=scenario,y=SA]{\srdata};

    \legend{MPC, DQN, DWA, SA-MPC}
    \end{axis}
  \end{tikzpicture}
    \vspace{-3mm}
    \caption{Success rate across four scenarios. Zero-height bars indicate 0\% success.}
    \vspace{-3mm}
  \label{fig:method_comparison}
\end{figure}
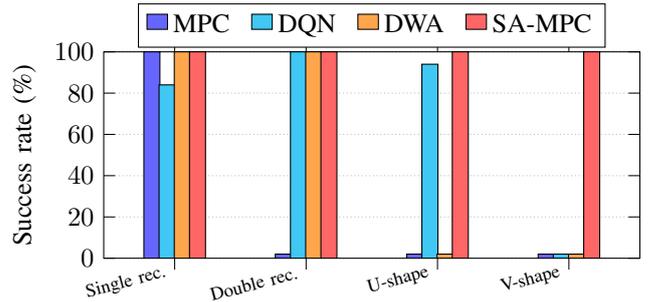

The framework was tested across five scenarios, as shown in Fig.~\ref{fig:mpc_scenarios}, ranging from simple convex obstacles to complex non-convex geometries. In the narrow dual-obstacle case, the controller generates smooth trajectories and accurately navigates through narrow corridors. In U-shaped and V-shaped non-convex scenarios, the safe-area decomposition helps the trajectory avoid getting trapped in local minima, which is a common failure point for traditional approaches. It also ensures that necessary deviations remain smooth. Finally, in the cluttered scenario with multiple obstacles of varying shapes, the proposed method maintains successful navigation and computational efficiency.
In contrast, Fig.~\ref{fig:ref_mpc_scenarios} shows the performance of the reference controller in complex environments, where the robot gets stuck at local minima. 

Table \ref{tab:comparison_results} provides a detailed comparison illustrating the performance advantages of the proposed approach. While baseline MPC succeeds only with convex obstacles, our method achieves 100\% success across all scenarios. For non-convex obstacles, baseline MPC exhibits computational spikes and fails to find solutions, whereas SA-MPC maintains consistent performance with very low maximum computation times, including the full pipeline overhead.

\subsection{Extensive Comparative Study}
Apart from the baseline MPC method, a learning-based method and a sampling-based method are also adopted for comparison. As demonstrated in Fig.~\ref{fig:method_comparison}, success rates across four scenarios are compared against baseline MPC, learning-based Deep Q-Network (DQN) method, and sampling-based Dynamic Window Approach (DWA).
SA-MPC achieves 100\% success across all scenarios. Baseline MPC handles only simple convex cases, failing in complex geometries where the solver encounters infeasibility. DQN shows moderate performance but struggles with precise manoeuvring, which causes hesitation and collision when encountering obstacles with sharp corners \cite{drlmpc_paper}. DWA excels in simple scenarios but cannot handle non-convex obstacles, since it has limited flexibility for necessary detours. SA-MPC's superior performance stems from transforming non-convex collision avoidance into convex constraints via quadtree-derived safe areas, guaranteeing recursive feasibility.

\begin{table*}[!t]
  \centering
  \caption{Performance comparison between Safe‑Area (SA) MPC and baseline MPC approaches (averaged over 10 runs)}
  \label{tab:comparison_results}
  \resizebox{\textwidth}{!}{%
  \begin{tabular}{@{}llcccccc@{}}
    \toprule
    \textbf{Scene} & \textbf{Obstacle Type} & \textbf{Method}
      & \multicolumn{3}{c}{\textbf{MPC Computation / Full Pipeline* (ms/step)}}
      & \multicolumn{2}{c}{\textbf{Performance}} \\ 
    \cmidrule(lr){4-6} \cmidrule(lr){7-8}
                   &                        & 
      & \textbf{Mean} & \textbf{Max} & \textbf{Std}
      & \textbf{Steps} & \textbf{Success (\%)} \\ 
    \midrule
    \multirow{8}{*}{Scene\,1} 
      & \multirow{2}{*}{\makecell[l]{Single rectangular\\obstacle}} % small
      & Baseline‑MPC & 23.7   & 273.5  & 38.8  & 66   & 100 \\
      &   & \textbf{SA‑MPC}   & \textbf{20.34} / 29.18  & \textbf{39.56} / 52.67  & \textbf{8.19} / 12.24 & 58   & \textbf{100} \\
      
    \cmidrule(lr){2-8}
      & \multirow{2}{*}{\makecell[l]{Two rectangular\\obstacles}} % large
          & Baseline‑MPC & 24.5   & 75     & 17.9  & 71   & 100 \\
        &   & \textbf{SA‑MPC}   & \textbf{22.6} / 46.83   & \textbf{87.04} / 108.4  & \textbf{13.67} / 22.46 & 83     & \textbf{100} \\
    \cmidrule(lr){2-8}
      & \multirow{2}{*}{\makecell[l]{U‑shape obstacle}} % large
        & Baseline‑MPC & 126.2  & 949.4  & 11.75 & -   & 0 \\
      &   & \textbf{SA‑MPC}   & \textbf{23.79} / 38.4  & \textbf{59.77} / 88.76  & \textbf{10.93} / 18.73 & 62 & \textbf{100}   \\
    \cmidrule(lr){2-8}
      & \multirow{2}{*}{\makecell[l]{V‑shape obstacle}} % large
        & Baseline‑MPC & 97.8   & 694.5  & 196.2 & -     & 0 \\
      &   & \textbf{SA‑MPC}   & \textbf{22.8} / 38.4   & \textbf{55.41} / 85.83  & \textbf{11.6} / 21.85 & 64 & \textbf{100}   \\
    \midrule
    \multirow{2}{*}{Scene\,2} 
      & \multirow{2}{*}{\makecell[l]{Complex map\\(mixed obstacles)}} 
        & Baseline‑MPC & 19.6   & 58.1   & 6.9  & -    & 0 \\
      &   & \textbf{SA‑MPC}   & \textbf{25.71} / 35.02  & \textbf{48} / 57.3     & \textbf{7} / 7.5 & 122    & \textbf{100} \\
    \bottomrule
    \multicolumn{8}{@{}l}{\small *Full pipeline (only for SA-MPC) includes quadtree generation, safe area merging, and trajectory planning.}
  \end{tabular}%
  }
\end{table*}

\section{Discussion}

The experimental results demonstrate that the proposed SA-MPC framework outperforms the baseline MPC approach across all tested scenarios, in terms of both efficiency and safety, particularly in environments with complex non-convex obstacles. Its core innovation is using convex collision-free regions from a hierarchical quadtree decomposition both to generate trajectories and to impose soft constraints within the MPC. This transforms non-convex obstacle avoidance into a structured, efficient pipeline that avoids explicit obstacle modelling, unlike classical MPC methods that encode obstacles as non-linear constraints, signed-distance penalties, or discrete exclusion zones.
The generation of collision-free areas guarantees recursive feasibility at each step, enabling MPC to compute valid control actions even in non-convex scenarios. 
As shown in Table~\ref{tab:comparison_results}, our system consistently produces collision-free, feasible trajectories with low optimisation latency. While a baseline MPC comparison is provided, it lacks integrated trajectory generation and collision-free-area constraints. Therefore, these results underscore the enhanced capabilities of the proposed integration approach.

In the current development, the proposal framework has a structural limitation, which is the reliance on axis-aligned rectangular safe areas. This may introduce conservatism in environments where free-space boundaries are highly irregular or diagonally oriented, and result in suboptimal paths or reduced manoeuvrability. Furthermore, the potential extension to dynamic settings will require efficient incremental updates of the quadtree and its connectivity graph, as well as providing real-time guarantees, which is a challenging task and an important future work.

\section{Conclusion}
This work presents an integrated navigation framework that unifies perception, environment representation, trajectory planning, and MPC through a quadtree-based safe-area decomposition. By converting obstacle avoidance into convex constraints, it ensures recursive feasibility, efficiency, and robustness without heuristic penalties. The direct integration of quadtree-derived safe areas into the MPC controller simplifies obstacle handling, enhances interpretability, and maintains strong performance across complex non-convex environments—offering a fast, reliable, and verifiable navigation pipeline for real-world deployment.

Future work will explore dynamic updates to the quadtree to handle moving obstacles, as well as extensions to multi-robot coordination. Real-world deployment will also be pursued to validate robustness under practical constraints.

\addtolength{\textheight}{-12cm} 

%%%%%%%%%%%%%%%%%%%%%%%%%%%%%%%%%%%%%%%%%%%%%%%%%%%%%%%%%%%%%%%%%%%%%%%%%%%%%%%%

\bibliographystyle{IEEEtran}
\bibliography{main}

\end{document}